\documentclass{article}
\usepackage{iclr2027_conference,times}

\usepackage{amsmath,amsfonts,bm}

\def\eqref#1{equation~\ref{#1}}

\def\1{\bm{1}}

\DeclareMathAlphabet{\mathsfit}{\encodingdefault}{\sfdefault}{m}{sl}
\SetMathAlphabet{\mathsfit}{bold}{\encodingdefault}{\sfdefault}{bx}{n}

\usepackage{hyperref}
\usepackage{url}
\usepackage{graphicx}
\usepackage{booktabs}
\usepackage{multirow}
\usepackage{xcolor}
\usepackage{amsmath}
\usepackage{amssymb}
\usepackage{caption}
\usepackage{placeins}
\usepackage{float}

\title{How to Reduce Localization Ambiguity? Geometry-Semantic Constrained BEV Representation Learning for Satellite-Ground Localization}
\author{
\normalfont
\begin{minipage}{0.96\textwidth}
\centering
\textbf{Junming Feng}$^{1,2,\dagger}$, \textbf{Panwang Xia}$^{3}$, \textbf{Qiong Wu}$^{3}$,
\textbf{Xudong Lu}$^{4}$, \textbf{Zeyu Jiao}$^{5}$, \textbf{Kun Lv}$^{5}$,\\
\textbf{Zherong Wu}$^{4}$, \textbf{Yi Wan}$^{3}$, \textbf{Peifeng Ma}$^{4}$,
\textbf{Li-Ta Hsu}$^{1}$, \textbf{Zhi Zheng}$^{1,*}$\\[0.55em]
{\small
$^{1}$The Hong Kong Polytechnic University, Hong Kong\\
$^{2}$Southern University of Science and Technology\\
$^{3}$Wuhan University, Wuhan, China\\
$^{4}$The Chinese University of Hong Kong, Hong Kong, China\\
$^{5}$Huawei Technologies Co., Ltd\\[0.55em]
}
{\scriptsize\ttfamily
fengjm2023@mail.sustech.edu.cn; xiapanwang@whu.edu.cn; mabel\_wq@whu.edu.cn\\
luxudong@link.cuhk.edu.hk; jiaozeyu2@huawei.com; lvkun@huawei.com\\
zherongwu@cuhk.edu.hk; mapeifeng@cuhk.edu.hk\\
lt.hsu@polyu.edu.hk; zhi.zheng@polyu.edu.hk
}
\end{minipage}
}

\iclrfinalcopy

\begin{document}
\maketitle
\lhead{}
\ificlrfinal
\begingroup
\renewcommand{\thefootnote}{\fnsymbol{footnote}}
\footnotetext[1]{Corresponding author: Zhi Zheng (zhi.zheng@polyu.edu.hk).}
\footnotetext[2]{This work was carried out while Junming Feng was a visiting student at The Hong Kong Polytechnic University.}
\endgroup
\fi

\begin{abstract}
  Satellite-ground localization estimates the planar position and yaw orientation of a ground camera within a geo-referenced satellite image of its surroundings.
  The predominant approach to this task maps features from ground-view images and satellite references into a shared bird's-eye-view (BEV) space and then establishes spatial correspondences between them.
  Although effective, this approach still faces ambiguity in feature placement and descriptor matching.
  When mapping ground-view features into BEV space, insufficient depth constraints allow the same feature to be assigned to different distances along the viewing direction, creating geometric ambiguity in BEV feature placement.
  Meanwhile, similar appearances at different locations create descriptor matching ambiguity, and existing descriptor learning lacks explicit semantic supervision to distinguish these locations.
  To reduce these ambiguities, we propose GeoSem-BEV, a geometry-semantic constrained BEV representation learning method.
  First, we geometrically constrain ground-view BEV feature placement through radial depth supervision for distance assignment and vertical height supervision for height aggregation.
  Then, shared explicit semantic supervision further promotes consistent semantic predictions across views and helps distinguish different locations with similar semantics.
  These constraints jointly improve feature placement and descriptor discriminability, thus improving the satellite-ground localization performance of state-of-the-art models by a large margin.
  Qualitative and quantitative results demonstrate the effectiveness of GeoSem-BEV in enhancing these models.
  On VIGOR with unknown orientation, GeoSem-BEV reduces mean orientation error relative to the corresponding state-of-the-art method by 37.2\% and 38.1\% in the cross-area and same-area settings, respectively.
  The corresponding errors were reduced by 10.8\% and 15.6\% on DReSS-D.
  On KITTI-CVL, GeoSem-BEV reduces the same-area mean orientation error by 26.8\% relative to the corresponding baseline under $\pm10^\circ$ orientation noise.

\end{abstract}

\section{Introduction}
\begin{figure}[t]
\centering
\includegraphics[width=\linewidth]{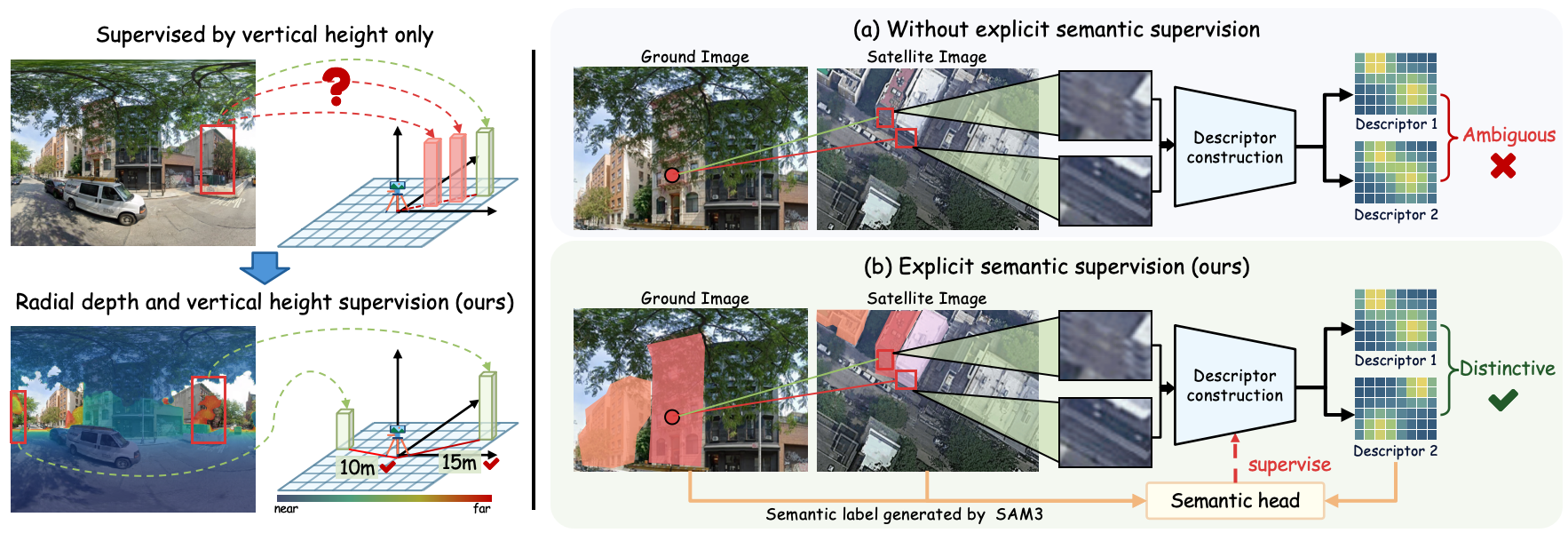}
\caption{Two ambiguities in BEV-space feature matching.
Left: Insufficient radial-depth constraints leave BEV feature placement ambiguous along viewing rays.
(a) Upper right: Similar appearances at different locations can produce ambiguous descriptors without explicit semantic supervision.
(b) Lower right: Explicit semantic supervision helps distinguish them.}
\label{fig:intro}
\end{figure}

Satellite-ground localization estimates the planar position and yaw orientation of a ground camera within a geo-referenced satellite image, supporting applications in vehicle localization and navigation, aerial-platform geolocation, and remote-sensing image registration for urban mapping and monitoring~\citep{hu2020image,shi2022beyond,zheng2020university,shan2014accurate,yu2021multimodal}.
Given a coarse Global Navigation Satellite System (GNSS) estimate of the surrounding area, satellite-ground localization can match ground observations with corresponding content in a satellite image to further refine the camera pose~\citep{shi2022beyond}.
However, substantial appearance differences between street-level ground images and overhead satellite observations make accurate spatial correspondence estimation challenging.

In recent years, satellite-ground localization has received increasing attention, and mapping ground-view features into a shared bird's-eye-view (BEV) space has become a predominant approach.
This shared top-down coordinate system reduces the viewpoint gap between ground and satellite views and makes their spatial correspondences explicit.
Existing methods construct and match BEV representations through geometric projection, learned feature transformation, displacement fields, or dense 3D feature sampling~\citep{shi2023boosting,fervers2023uncertainty,song2023learning,wang2023fine,xia2025fg2}.
Despite their effectiveness, these methods share a common challenge: the construction of the ground-view BEV representation remains insufficiently constrained along viewing rays.
Loc$^2$ highlights this issue by observing that BEV representation construction can introduce ray-directional distortions and lose height information.
It therefore matches features directly in the image planes, which we refer to as image-space feature matching, before lifting matched points using depth.
This pipeline requires a monocular depth prediction model at inference to determine the spatial coordinates of matched points~\citep{xia2026loc}.
Unlike Loc$^2$, ViewBridge continues to improve BEV representation construction through visible-surface modelling and height consistency constraints~\citep{xia2025viewbridge}.
Its supervision constrains vertical height, specifying where scene content lies along the vertical axis.
However, it ignores the ambiguity in radial depth, which determines how far that content lies from the camera along the viewing ray.
Consequently, the same image feature may contribute to BEV locations at different distances along the viewing ray, leaving geometric ambiguity in feature placement, as illustrated in the upper-left part of Fig.~\ref{fig:intro}.

Besides geometric ambiguity in feature placement, similar appearances at different locations can create descriptor matching ambiguity.
For example, visually similar regions at different locations may produce similar descriptors, allowing incorrect locations to receive high matching scores, as illustrated in Fig.~\ref{fig:intro}(a).
Meanwhile, drastic viewpoint differences can make the same location appear visually different in ground and satellite images, causing correct correspondences to receive low matching scores.
These appearance differences make semantic cues useful for associating corresponding locations.
Prior work uses semantic layouts for ground-to-GIS matching~\citep{castaldo2015semantic} or semantic map elements for localization~\citep{sarlin2023orienternet}.
Because these cues come from external maps, map errors can add uncertainty to cross-view matching~\citep{castaldo2015semantic}.
Within BEV matching, existing methods learn descriptors mainly from appearance, matching, and pose supervision, without explicit semantic supervision~\citep{xia2025fg2,xia2025viewbridge}.

To reduce these ambiguities, we propose GeoSem-BEV, a geometry-semantic constrained BEV representation learning method that jointly constrains feature placement and descriptor learning within BEV-space feature matching.
First, we impose geometric constraints on how ground-view features are assigned to and aggregated on the BEV grid.
Radial depth supervision constrains their distance from the camera along viewing rays, while vertical height supervision constrains the aggregation of features sampled at different heights.
We then introduce shared explicit semantic supervision for ground- and satellite-view descriptors.
SAM3~\citep{carion2026sam} generates semantic masks for paved surfaces, buildings, and vegetation as training targets.
A shared semantic head predicts semantic classes from ground- and satellite-view descriptors, allowing these targets to supervise descriptor learning in both views, as illustrated in Fig.~\ref{fig:intro}(b).
This semantic supervision promotes consistent semantic predictions at corresponding locations.
To further distinguish spatially different locations with similar semantics, we additionally select same-class hard negatives outside the correct spatial neighbourhood and penalize their high matching scores.
Together, shared semantic supervision and same-class hard negatives improve descriptor discriminability, while geometric constraints improve feature placement.
We evaluate GeoSem-BEV across ViewBridge~\citep{xia2025viewbridge}, FG$^2$~\citep{xia2025fg2}, and DenseFlow~\citep{song2023learning} on VIGOR~\citep{zhu2021vigor}, KITTI-CVL~\citep{shi2022cvlnet}, and DReSS-D~\citep{xia2025cross} to assess whether the proposed constraints improve localization across BEV matching frameworks.
On VIGOR with unknown orientation, GeoSem-BEV reduces mean orientation error relative to the corresponding state-of-the-art method by 37.2\% and 38.1\% in the cross-area and same-area settings, respectively.
The corresponding errors were also reduced by 10.8\% and 15.6\% on DReSS-D.
These results demonstrate that GeoSem-BEV improves satellite-ground localization accuracy across different matching frameworks while providing more discriminative correspondence evidence.

Our main contributions are as follows: \textbf{(i)} We analyse two sources of localization ambiguity in BEV-space feature matching: geometric ambiguity caused by insufficient constraints on feature placement along viewing rays, and descriptor matching ambiguity arising from similar appearances at different locations.
\textbf{(ii)} We propose GeoSem-BEV, a geometry-semantic constrained BEV representation learning method.
Geometric constraints guide BEV feature placement through radial depth and vertical height supervision.
Shared explicit semantic supervision promotes cross-view semantic consistency, while same-class hard negatives further improve descriptor discriminability.
\textbf{(iii)} We evaluate GeoSem-BEV by applying it to three state-of-the-art BEV localization models, ViewBridge, FG$^2$, and DenseFlow, on VIGOR, KITTI-CVL~\citep{shi2022cvlnet}, and DReSS-D.
On VIGOR with unknown orientation, GeoSem-BEV reduces mean orientation error relative to the corresponding state-of-the-art method by 37.2\% and 38.1\% in the cross-area and same-area settings, respectively, and achieves the lowest mean translation error among compared methods in both settings.
On DReSS-D and KITTI-CVL, it also achieves competitive results, improving orientation and localization metrics over the corresponding baseline models.

\section{Related Work}

\subsection{Satellite-Ground Localization}

Satellite-ground localization has progressed from retrieving a matching reference image to estimating the pose of a ground camera within that image.
Early cross-view methods mainly addressed ground-to-aerial image retrieval by learning shared representations~\citep{lin2013cross,lin2015learning,workman2015wide}.
Subsequent retrieval models improved cross-view representation learning with convolutional global descriptors~\citep{hu2018cvm} and transformer-based spatial context~\citep{zhu2022transgeo}.
Later studies moved beyond image retrieval to estimate the position and orientation of a ground camera from overhead imagery~\citep{vo2016localizing,shi2022beyond}.
VIGOR relaxed the one-to-one retrieval assumption by evaluating camera locations beyond satellite-image centres~\citep{zhu2021vigor}, while subsequent methods explored differentiable geometric alignment~\citep{shi2022beyond}, pose-dependent descriptor comparison~\citep{lentsch2023slicematch}, and convolutional cross-view pose estimation~\citep{xia2023convolutional}.
These developments shifted the focus from image retrieval toward fine-grained pose estimation and explicit spatial correspondence.
They also support applications such as vehicle localization with temporal filtering~\citep{hu2020image} and drone-view target localization and navigation~\citep{zheng2020university}.
For fine-grained pose estimation, mapping ground-view and satellite features into a shared BEV space has become a major approach.

Within BEV-space feature matching, earlier methods estimate pose distributions~\citep{fervers2023uncertainty} or homographies from cross-view correlations~\citep{wang2023fine}.
Correspondence-based approaches make spatial associations explicit and recover pose from matched coordinates.
DenseFlow predicts displacement fields between the two views~\citep{song2023learning}, while FG$^2$ matches BEV descriptors and samples correspondences for geometric alignment~\citep{xia2025fg2}.
ViewBridge further improves BEV representation construction through visible-surface modelling and matching-score refinement~\citep{xia2025viewbridge}.
Loc$^2$ identifies distortions and information loss introduced during BEV transformation and explores an image-space feature matching route, which matches features in the image planes before lifting the matched points using depth~\citep{xia2026loc}.
These studies motivate further investigation of the BEV-space feature matching route, where the construction of ground-view BEV representations and the learning of matching descriptors remain subject to ambiguity.
Our work focuses on reducing geometric ambiguity in feature placement and descriptor matching ambiguity within this BEV-based framework.

\subsection{Geometric Constraints for BEV Representation Construction}

BEV representation construction maps image features to a top-down grid, with geometric constraints guiding their spatial placement.
Related work in autonomous-driving perception explores depth-based lifting and attention-based feature aggregation.
Lift-Splat-Shoot distributes image features over depth hypotheses before aggregating them on a BEV grid~\citep{philion2020lift}.
BEVDepth further introduces explicit depth supervision to improve this transformation for 3D object detection~\citep{li2023bevdepth}.
BEVFormer uses spatial cross-attention and temporal self-attention to construct BEV representations for 3D detection and map segmentation~\citep{li2024bevformer}.
Although these methods target multi-camera perception, they provide two relevant principles for BEV construction: assigning image features to spatial locations and aggregating features over the BEV grid.

Satellite-ground localization applies similar construction operations under a different camera geometry.
DenseFlow projects ground features under a fixed camera-height assumption, while GGCVT combines geometric projection with learned attention~\citep{song2023learning,shi2023boosting}.
FG$^2$ samples ground-view features at 3D queries and learns to aggregate them over height; ViewBridge further introduces visible-surface modelling and explicit height consistency constraints~\citep{xia2025fg2,xia2025viewbridge}.
These methods constrain how features are distributed along the vertical axis, yet they do not directly supervise radial depth along viewing rays.
Consequently, the same image feature may contribute to BEV locations at different distances from the camera, leaving geometric ambiguity in BEV feature placement.

\subsection{Semantic Information for Cross-View Matching}

Semantic information provides cues beyond visual appearance for associating corresponding content across viewpoints.
Semantic Cross-View Matching uses semantic categories and their spatial layout to associate street-level observations with GIS maps~\citep{castaldo2015semantic}.
OrienterNet aligns image-derived BEV features with semantic elements in public maps for image localization~\citep{sarlin2023orienternet}.
SNAP learns neural maps from ground and overhead imagery, with semantic structure emerging during map learning without explicit semantic labels~\citep{sarlin2023snap}.
For cross-view remote-sensing applications, SG-BEV further combines satellite and ground observations for building-attribute segmentation~\citep{ye2024sg}.
These studies demonstrate that semantic information can complement appearance and geometry in cross-view association.
However, methods relying on external semantic maps may inherit errors, missing entries, or temporal changes in the reference-side annotations, which can introduce uncertainty into matching~\citep{castaldo2015semantic}.

For local BEV feature matching, semantic information is also relevant to descriptor ambiguity caused by similar appearances at different locations.
Prior work has studied this issue in joint cross-view retrieval and offset calibration by modelling semantic ambiguity through offset uncertainty~\citep{feng2025semantic}.
General descriptor-learning methods use supervised contrastive objectives or hard-negative mining to improve class-level discrimination~\citep{khosla2020supervised,mishchuk2017working}.
However, these objectives alone do not establish spatial correspondence, since different buildings or road segments may share the same semantic class.
Existing BEV matching methods primarily learn descriptors through appearance, matching, and pose supervision, without explicitly supervising the semantic content encoded by the descriptors.
This motivates explicit semantic supervision for ground- and satellite-view descriptors, which provides scene-content information beyond appearance while preserving the distinction between spatially different locations.

\section{Method}

\subsection{Overview}

Given a ground-view image $G$ and a geo-referenced satellite image $S$ covering its surroundings, satellite-ground localization estimates the planar camera pose $\hat T=(\hat t_x,\hat t_y,\hat\theta)$, where $\hat t_x$ and $\hat t_y$ denote translation in the satellite reference frame and $\hat\theta$ denotes yaw orientation.
Let $T^\ast$ denote the corresponding ground-truth pose.
The pose is recovered from cross-view correspondences between the ground observation and the satellite reference.
In the BEV-space feature matching route, ground-view features are first assigned to a top-down grid and then matched with satellite features in the shared coordinate system.
However, insufficient constraints on this assignment can create radial placement ambiguity, while insufficient descriptor discrimination can favour visually similar but spatially incorrect matches.
GeoSem-BEV addresses these limitations by constraining both BEV feature placement and descriptor learning. The method is designed as a framework-agnostic enhancement: each evaluated model retains its original correspondence estimator, pose solver, and base localization objective.
GeoSem-BEV adds supervision at two complementary stages of the shared pipeline.
Geometry supervision acts during BEV representation construction, constraining ground-view feature placement on the BEV grid.
Semantic supervision acts on the resulting descriptor fields, encouraging descriptors from corresponding ground and satellite locations to encode compatible scene content while separating spatially different candidates. 

\begin{figure}[t]
  \centering
  \includegraphics[width=\linewidth]{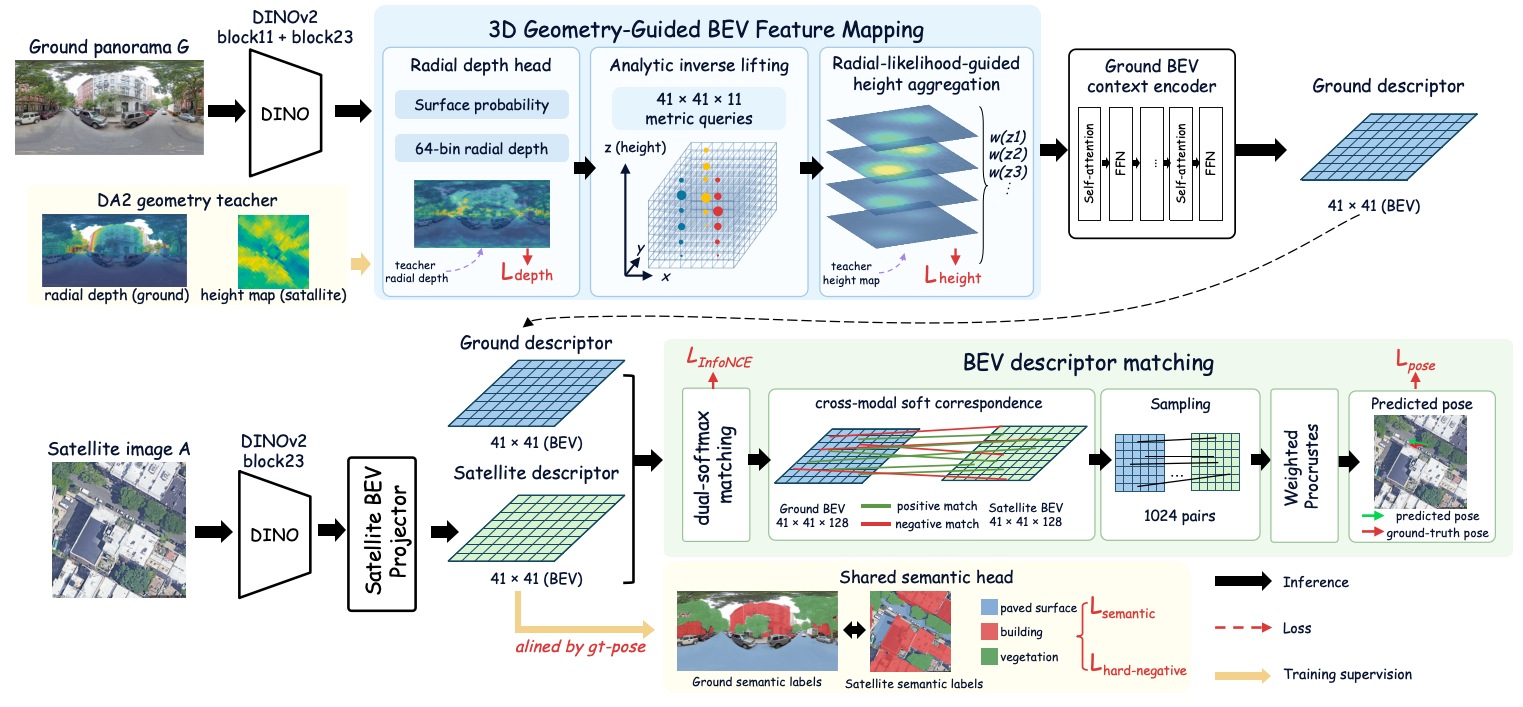}
  \caption{GeoSem-BEV with FG$^2$ as the base model.
  Geometric supervision constrains radial assignment and height aggregation of ground-view features.
  A shared semantic head supervises both descriptor fields, and same-class hard negatives penalize spatially incorrect matches.}
  \label{fig:method}
\end{figure}

Figure~\ref{fig:method} illustrates GeoSem-BEV with FG$^2$ as the base model~\citep{xia2025fg2}.
The following formulation describes the geometric and semantic constraints used across the evaluated BEV matching frameworks.
The pipeline extracts ground-view and satellite-view features, samples ground features at metric 3D queries, and aggregates them into a 2D BEV grid.
Radial-depth and vertical-height supervision constrain feature placement and height aggregation, after which projection heads produce descriptors for correspondence estimation.
During training, SAM3-derived targets supervise a shared semantic head, while pose-aligned consistency and same-class hard negatives improve descriptor discrimination.
At inference, only the ground and satellite images are required.
Framework-specific implementations are described in Appendix~\ref{app:pipeline-adaptations}.

\subsection{Geometry-Constrained BEV Feature Construction}

The geometric constraint addresses the construction of a ground-view BEV representation before correspondence estimation.
For each image feature, the viewing ray determines a direction, but the feature can still be assigned to multiple distances along that ray.
We therefore distinguish radial depth from vertical height: radial depth specifies the distance assignment along a viewing ray, whereas vertical height specifies how samples at different heights contribute to one BEV cell.
Radial-depth supervision and vertical-height supervision jointly constrain these two parts of BEV feature placement.
\paragraph{Sampling ground-view features.}
Let $q_{ijk}=(x_i,y_j,z_k)$ be a fixed metric 3D query in the vertical column associated with BEV cell $(x_i,y_j)$.
The ground coordinate frame is centred at the camera, with two horizontal axes and a vertical axis.
The camera projection $\pi(q_{ijk})$ identifies the image location from which a feature is sampled:
\begin{equation}
 F_{3D}(q_{ijk})=\operatorname{sample}(F_G,\pi(q_{ijk})).
 \label{eq:inverse-lifting}
\end{equation}
Each BEV cell receives candidate appearance features at the queried heights.
Sampling establishes a viewing direction for each candidate, but does not determine whether a surface exists at that distance.

\paragraph{Radial assignment and height aggregation.}
A depth head predicts a surface probability $p_s(u,v)$ and a discrete conditional distribution $p_r(b\mid u,v)$ over radial depth bins with centres $d_b$.
Let $\widetilde p_r(r\mid u,v)$ denote its linearly interpolated value at distance $r$.
For a query at radial distance $r_{ijk}=\lVert q_{ijk}\rVert_2$, its radial likelihood is
\begin{equation}
 \ell_{ijk}=p_s(\pi(q_{ijk}))\,\widetilde p_r(r_{ijk}\mid\pi(q_{ijk})).
 \label{eq:radial-weight}
\end{equation}
Queries beyond the supported depth range receive zero likelihood.
This distinguishes candidate locations along the same viewing ray using the predicted surface distance.
The radial likelihood is combined with the height aggregation weights to obtain the contribution weight $w_{ijk}$.
The ground BEV feature is
\begin{equation}
 F_{\mathrm{BEV}}(x_i,y_j)=\sum_k w_{ijk}F_{3D}(q_{ijk}).
 \label{eq:height-aggregation}
\end{equation}
Radial predictions therefore constrain where sampled ground-view features contribute to the BEV grid.
The aggregation weights also receive vertical-height supervision, as described below.

\paragraph{Geometry supervision.}
Frozen monocular depth teachers provide pseudo-targets for radial distance and relative height during training.
Ground-view targets supervise surface presence, radial depth, expected depth, and uncertainty, while the ground-truth pose $T^\ast$ aligns satellite height targets with the ground BEV grid.
The geometry objective combines these terms:
\begin{equation}
 \mathcal L_{\mathrm{geom}}=\mathcal L_{\mathrm{surface}}+
 \mathcal L_{\mathrm{radial\text{-}dist}}+\mathcal L_{\mathrm{depth}}+
 \mathcal L_{\mathrm{uncertainty}}+\mathcal L_{\mathrm{height}}.
 \label{eq:geometry-loss}
\end{equation}
At inference, the image-based student predicts the geometric weights without running the teacher.
Thus, the teacher supplies training-time geometric targets, while the student produces the constrained BEV representation from the input images alone at inference.

\subsection{Explicit Semantic Supervision for BEV Descriptors}

After geometric supervision constrains ground-view feature placement, semantic supervision constrains descriptor content.
Matching and pose supervision alone can still assign high scores to visually similar regions at different locations.

\paragraph{Semantic targets and shared prediction.}
The ground BEV feature map and satellite feature map are encoded into descriptor fields $D_G$ and $D_S$.
We use SAM3~\citep{carion2026sam} to generate masks for paved surfaces, buildings, and vegetation in both images.
Ground and satellite semantic masks are converted into soft targets on their respective BEV grids.
The resulting soft class targets $Y_G$ and $Y_S$ are fixed supervision signals, distinct from the student's predicted class distributions.

A shared head $h$ predicts class probabilities $P_v=\operatorname{softmax}(h(D_v))$, for $v\in\{G,S\}$.
The soft cross-entropy loss $\mathcal L_{\mathrm{CE}}^v$ compares these predictions with the SAM3-derived targets $Y_v$ over valid BEV cells.
Gradients through the shared head teach the matching descriptors to encode semantics together with appearance.

We warp $P_G$ with the ground-truth pose and enforce Jensen--Shannon consistency with $P_S$ at corresponding locations:
\begin{equation}
 \mathcal L_{\mathrm{semantic}}=
 \tfrac12(\mathcal L_{\mathrm{CE}}^G+\mathcal L_{\mathrm{CE}}^S)
 +\mathcal L_{\mathrm{JS}}(P_G^{T^\ast},P_S).
 \label{eq:semantic-loss}
\end{equation}
Here, $P_G^{T^\ast}$ denotes the warped ground-view class probabilities.

\paragraph{Distinguishing same-class locations.}
Same-class hard negatives are selected outside the positive spatial neighbourhood to distinguish locations with similar semantics.
The directional hard-negative loss contrasts the positive correspondence with these spatially incorrect candidates using the matching logits.
The reverse direction is defined analogously, and we average both directions:
\begin{equation}
 \mathcal L_{\mathrm{HN}}=\tfrac12
 (\mathcal L_{G\rightarrow S}^{\mathrm{HN}}+
 \mathcal L_{S\rightarrow G}^{\mathrm{HN}}).
 \label{eq:hard-negative-loss}
\end{equation}
Together, these terms align descriptors of corresponding content and distinguish visually similar regions at different locations.
The semantic head and SAM3 targets are used only during training.

\subsection{Correspondence Sampling and Training}

The descriptor fields retain their BEV coordinates and are processed by each framework's correspondence estimator, which produces candidate matches with nonnegative confidence weights after the proposed geometric and semantic supervision.
This separation keeps the proposed representation-learning constraints independent of the implementation-specific form of correspondence estimation.

The selected correspondences provide paired metric BEV coordinates $g_n$ and $s_n$, with nonnegative confidence weights $c_n$.
Ground coordinates are relative to the camera, and satellite coordinates are relative to the reference image centre.
The evaluated pipelines recover the planar pose by minimizing the weighted alignment error:
\begin{equation}
 \hat T=\underset{T\in\mathrm{SE}(2)}{\arg\min}
 \sum_{n=1}^{N}c_n\lVert s_n-T(g_n)\rVert_2^2.
 \label{eq:procrustes}
\end{equation}
Here, $\mathrm{SE}(2)$ denotes planar rigid transformations consisting of two-dimensional translation and yaw rotation.
This objective uses the rigid-alignment form of Procrustes analysis~\citep{umeyama1991least}.

Training retains each pipeline's localization and matching objective $\mathcal L_{\mathrm{base}}$ and adds the proposed supervision:
\begin{equation}
 \mathcal L=\mathcal L_{\mathrm{base}}+
 \lambda_g\mathcal L_{\mathrm{geom}}+
 \lambda_s\mathcal L_{\mathrm{semantic}}+
 \lambda_h\mathcal L_{\mathrm{HN}}.
 \label{eq:total-loss}
\end{equation}
The coefficients $\lambda_g$, $\lambda_s$, and $\lambda_h$ balance geometric supervision, semantic supervision, and same-class hard-negative learning, respectively.
The base localization and matching objective is retained for each framework, while the three additional terms provide the proposed supervision.

\section{Experiments}
\label{sec:experiments}

\subsection{Datasets and Evaluation Metrics}

\paragraph{Datasets and protocols.}
We evaluate on VIGOR~\citep{zhu2021vigor}, KITTI-CVL~\citep{shi2022cvlnet}, and DReSS-D~\citep{xia2025cross}.
VIGOR contains ground panoramas and aerial images from four US cities, with same-area and cross-area splits.
The comparison reports unknown-orientation results for both cross-area and same-area splits.
DReSS-D evaluates decentrality, where the ground camera need not lie at the centre of its satellite reference tile.
Its same-area and cross-area unknown-orientation results are provided in Appendix~\ref{app:dressd-results}.
KITTI-CVL is a limited-field-of-view satellite-ground localization benchmark built from KITTI driving imagery and satellite references.
The complete KITTI-CVL protocol and results are provided in Appendix~\ref{app:kitti-results}.

\paragraph{Metrics.}
We report mean and median translation error in metres and orientation error in degrees.
VIGOR additionally uses joint localization recall at $(1\,\mathrm m,5^\circ)$, $(3\,\mathrm m,10^\circ)$, and $(5\,\mathrm m,20^\circ)$.
KITTI-CVL reports orientation recall within $1^\circ$ and $5^\circ$.
Recall is expressed as a percentage; lower errors and higher recalls indicate better performance.

\subsection{Implementation Details}

All evaluated variants use the same GeoSem-BEV supervision while retaining the correspondence and pose-estimation components of their underlying frameworks.
We refer to the GeoSem-BEV implementations of ViewBridge, FG$^2$, and DenseFlow as GS-ViewBridge, GS-FG$^2$, and GS-DenseFlow, respectively.
Training and evaluation configurations are summarized in Appendix~\ref{app:training-configuration}.

The matching coefficient is $\beta=100$ for unknown orientation and $\beta=1$ for known orientation.
Sem-FG$^2$ adds semantic supervision with uniform height aggregation, Geo-FG$^2$ adds geometric supervision, and GS-FG$^2$ combines both with same-class hard negatives.
The same geometric and semantic constraints are incorporated into DenseFlow and ViewBridge while retaining their original correspondence and pose-estimation components.
Depth predictions and SAM3 masks are used only to construct training targets; inference uses only the ground and satellite images.

\subsection{Quantitative Results}

Across the reported datasets, GeoSem-BEV consistently improves the most challenging unknown-orientation evaluations.
On VIGOR, all three implementations reduce both mean localization and orientation errors in the cross-area and same-area settings, with the largest overall gains obtained by GS-ViewBridge and GS-FG$^2$.
On DReSS-D, GS-FG$^2$ reduces both errors in the reported same-area and cross-area settings, while GS-ViewBridge provides its clearest gains in orientation estimation (Appendix~\ref{app:dressd-results}).
On KITTI-CVL, the limited-field-of-view evaluation shows comparable improvements in orientation and selected localization measures for the enhanced FG$^2$ and ViewBridge models, supporting the transfer of GeoSem-BEV beyond panoramic ground observations.
Together, these results show that the proposed constraints improve different BEV matching frameworks, while the magnitude and type of the gain depend on the underlying representation and evaluation setting.

\textbf{VIGOR.}
Table~\ref{tab:vigor-main} compares the GeoSem-BEV implementations of DenseFlow, FG$^2$, and ViewBridge with SliceMatch~\citep{lentsch2023slicematch}, CCVPE~\citep{xia2023convolutional}, DenseFlow~\citep{song2023learning}, FG$^2$, Loc$^2$, and ViewBridge under unknown orientation.
Published baselines follow the VIGOR comparisons reported in Loc$^2$ and ViewBridge~\citep{xia2026loc,xia2025viewbridge}, while FG$^2$ uses its original single-stage results~\citep{xia2025fg2} and the GeoSem-BEV implementations are evaluated on the official unknown-orientation test splits.

\begin{table}[t]
\centering
\caption{VIGOR unknown-orientation test results.
Lower is better.
GeoSem-BEV implementations are shown in black bold; best, second-best, and third-best distinct values within each area split are marked in red bold, blue bold, and black bold, respectively.
Ties share the same rank.
FG$^2$ reports original single-stage results; published baselines follow the cited comparisons.
For each GeoSem-BEV implementation, percentages in parentheses after every metric report the relative change from its corresponding base model, computed as $(E_{\mathrm{GS}}-E_{\mathrm{base}})/E_{\mathrm{base}}\times100\%$.}
\label{tab:vigor-main}
\resizebox{\linewidth}{!}{%
\begin{tabular}{lcccccccc}
\toprule
 & \multicolumn{4}{c}{Cross-area} & \multicolumn{4}{c}{Same-area}\\
\cmidrule(lr){2-5}\cmidrule(lr){6-9}
 & \multicolumn{2}{c}{Localization (m)$\downarrow$} & \multicolumn{2}{c}{Orientation ($^\circ$)$\downarrow$} & \multicolumn{2}{c}{Localization (m)$\downarrow$} & \multicolumn{2}{c}{Orientation ($^\circ$)$\downarrow$}\\
Method & Mean & Median & Mean & Median & Mean & Median & Mean & Median\\
\midrule
SliceMatch & 7.220 & 3.310 & 25.970 & 4.510 & 6.490 & 3.130 & 25.460 & 4.710 \\
CCVPE & 5.410 & \textcolor{red}{\textbf{1.890}} & 27.780 & 13.580 & \textcolor{black}{\textbf{3.740}} & \textcolor{red}{\textbf{1.420}} & 12.830 & 6.620 \\
DenseFlow & 7.670 & 3.670 & 17.630 & 2.940 & 4.970 & 1.900 & 11.200 & 1.590 \\
\textcolor{black}{\textbf{GS-DenseFlow}} & 5.647\,(-26.38\%) & 2.344\,(-36.13\%) & 15.466\,(-12.27\%) & 2.157\,(-26.63\%) & 4.677\,(-5.90\%) & 1.906\,(+0.32\%) & \textcolor{black}{\textbf{9.340}}\,(-16.61\%) & 1.589\,(-0.06\%)\\
FG$^2$ & 10.020 & 8.140 & 31.410 & 5.450 & 8.950 & 7.320 & 15.020 & 2.940 \\
\textcolor{black}{\textbf{GS-FG$^{\boldsymbol{2}}$}} & 4.662\,(-53.47\%) & 2.364\,(-70.96\%) & 13.850\,(-55.91\%) & \textcolor{blue}{\textbf{1.241}}\,(-77.23\%) & 3.823\,(-57.28\%) & 1.849\,(-74.74\%) & 9.723\,(-35.27\%) & \textcolor{blue}{\textbf{0.988}}\,(-66.39\%) \\
Loc$^2$ & \textcolor{black}{\textbf{4.230}} & 2.090 & \textcolor{black}{\textbf{11.670}} & 2.210 & 3.940 & \textcolor{black}{\textbf{1.780}} & 9.540 & 2.000 \\
ViewBridge & \textcolor{blue}{\textbf{3.790}} & \textcolor{blue}{\textbf{2.010}} & \textcolor{blue}{\textbf{10.590}} & \textcolor{black}{\textbf{2.040}} & \textcolor{blue}{\textbf{3.080}} & \textcolor{blue}{\textbf{1.570}} & \textcolor{blue}{\textbf{6.020}} & \textcolor{black}{\textbf{1.200}} \\
\textcolor{black}{\textbf{GS-ViewBridge}} & \textcolor{red}{\textbf{3.288}}\,(-13.25\%) & \textcolor{black}{\textbf{2.041}}\,(+1.54\%) & \textcolor{red}{\textbf{6.655}}\,(-37.16\%) & \textcolor{red}{\textbf{0.994}}\,(-51.27\%) & \textcolor{red}{\textbf{2.566}}\,(-16.69\%) & \textcolor{blue}{\textbf{1.570}}\,(+0.00\%) & \textcolor{red}{\textbf{3.724}}\,(-38.14\%) & \textcolor{red}{\textbf{0.784}}\,(-34.67\%) \\
\bottomrule
\end{tabular}}
\end{table}

On VIGOR with unknown orientation, all three implementations reduce mean localization and orientation errors relative to their corresponding base models.
GS-ViewBridge has the lowest mean localization and orientation errors in both area splits (Table~\ref{tab:vigor-main}).
Relative to DenseFlow, GS-DenseFlow reduces cross-area mean translation error from 7.670 m to 5.647 m and mean orientation error from 17.630$^\circ$ to 15.466$^\circ$, while GS-FG$^2$ reduces the corresponding errors from 10.020 m to 4.662 m and from 31.410$^\circ$ to 13.850$^\circ$ relative to FG$^2$.
The percentages in parentheses in Table~\ref{tab:vigor-main} report relative changes obtained by each GeoSem-BEV implementation over its base model.

\subsection{Ablation Study}

\begin{table}[t]
\centering
\caption{Ablation on VIGOR cross-area with unknown orientation.
All results use RANSAC.
Best values in each column are bold.
FG$^{2\dagger}$ denotes our reproduced FG$^2$ baseline.}
\label{tab:vigor-ablation}
\resizebox{\linewidth}{!}{%
\begin{tabular}{lccccccc}
\toprule
Method & Loc. mean $\downarrow$ & Loc. median $\downarrow$ & Ori. mean $\downarrow$ & Ori. median $\downarrow$ & R@1m/5$^\circ$ $\uparrow$ & R@3m/10$^\circ$ $\uparrow$ & R@5m/20$^\circ$ $\uparrow$\\
 & (m) & (m) & ($^\circ$) & ($^\circ$) & & & \\
\midrule
FG$^{2\dagger}$ & 5.751 & 2.943 & 18.908 & 1.337 & 13.07 & 50.11 & 67.04 \\
Sem-FG$^2$ & 6.339 & 3.335 & 22.740 & 1.397 & 11.06 & 45.25 & 62.76 \\
Geo-FG$^2$ & 5.014 & 2.469 & 15.980 & \textbf{1.240} & 15.57 & 57.75 & 74.26 \\
\textcolor{black}{\textbf{GS-FG$^{\boldsymbol{2}}$}} & \textbf{4.662} & \textbf{2.364} & \textbf{13.850} & 1.241 & \textbf{16.53} & \textbf{59.81} & \textbf{76.72} \\
\bottomrule
\end{tabular}}
\end{table}
Table~\ref{tab:vigor-ablation} compares geometry and semantic configurations on VIGOR cross-area with unknown orientation, using FG$^{2\dagger}$ as the baseline.
Sem-FG$^2$ uses semantic supervision with uniform height aggregation, Geo-FG$^2$ uses geometric supervision without semantic supervision, and GS-FG$^2$ combines both.
The ablation reveals an important interaction between the two constraints.
Adding semantic supervision alone in Sem-FG$^2$ degrades all reported metrics relative to the reproduced baseline, indicating that semantic supervision alone cannot compensate for insufficiently constrained BEV feature placement.
By contrast, Geo-FG$^2$ improves the baseline across the four pose-error metrics, showing that more accurate feature placement provides a stronger basis for correspondence learning.
Once this geometric constraint is present, adding semantic supervision and same-class hard negatives in GS-FG$^2$ further reduces the mean translation and orientation errors and yields the highest recalls.
These results suggest that semantic supervision is most effective after the BEV features have been placed at geometrically meaningful locations.

\FloatBarrier

\subsection{Discussion}

GeoSem-BEV shows its clearest gains in the more challenging VIGOR unknown-orientation evaluations, where all three adapted models reduce both mean localization and orientation errors.
The DReSS-D and KITTI-CVL results further show that the method transfers across datasets and limited-field-of-view observations, while the magnitude and type of improvement vary with the underlying framework and evaluation setting.
These differences suggest that geometric ambiguity in BEV feature placement and descriptor matching ambiguity are most consequential when the baseline retains substantial correspondence uncertainty; when the baseline already forms effective BEV features, the added constraints leave less ambiguity to resolve.
The ablation likewise shows that semantic supervision is most effective after geometric supervision has improved feature placement.

\section{Conclusion}

Satellite-ground localization estimates a ground-camera pose by matching ground-view and satellite-view observations across a large viewpoint difference.
BEV-space feature matching provides a common coordinate system, but underconstrained radial placement and insufficient descriptor discrimination can leave localization ambiguity.
We analyse these two sources of ambiguity and propose GeoSem-BEV to address them through complementary geometric and semantic supervision.
Specifically, radial-depth and vertical-height supervision constrain where ground-view features contribute on the BEV grid, reducing geometric ambiguity in feature placement.
Shared explicit semantic supervision promotes consistent semantic predictions across views, while same-class hard negatives improve descriptor discrimination and reduce descriptor matching ambiguity.
By incorporating GeoSem-BEV into state-of-the-art BEV localization models, including FG$^2$, DenseFlow, and ViewBridge, the resulting models achieve their most consistent improvements under unknown orientation on VIGOR, supporting the effectiveness of reducing localization ambiguity in BEV-space feature matching.
On VIGOR with unknown orientation, GeoSem-BEV reduces mean orientation error relative to the corresponding state-of-the-art method by 37.2\% and 38.1\% in the cross-area and same-area settings, respectively.
On DReSS-D, the corresponding mean orientation errors are reduced by 10.8\% and 15.6\% relative to the reproduced FG$^{2\dagger}$ baseline.
Taken together, the results show that GeoSem-BEV is most effective when the base model retains substantial geometric ambiguity in BEV feature placement or descriptor matching ambiguity.
Future work will improve GeoSem-BEV for limited-field-of-view settings, adapt its constraints across matching frameworks, and develop more accurate methods for BEV feature placement and cross-view matching.
\subsection*{Ethics Statement}

This work uses public cross-view localization benchmarks and does not collect new personal data or conduct experiments on human participants.
Dataset licenses, image-access restrictions, and the intended use of the released benchmarks should be followed when distributing code or derived artifacts.

\subsection*{AI Use Statement}

Large language models and other AI tools were used to assist with writing and language polishing, retrieve and identify related literature, support research execution through code development and engineering workflows, and draft manuscript sections from author-provided methods and experimental records.
The authors retain responsibility for the research decisions, experimental results, and final manuscript, including the accuracy and originality of the text.
AI-assisted literature suggestions and technical content are subject to author verification against original sources, implementation code, and experimental records.

\ificlrfinal
\subsubsection*{Acknowledgments}

The authors would like to thank Huawei for the support of Ascend NPUs, which provided the computing infrastructure for the experiments in this work.
We also gratefully acknowledge the open-source Ascend Agent Skills repository (\url{https://gitcode.com/Ascend/agent-skills}) for providing reusable agent skills and domain knowledge for the Ascend software stack, which facilitated AI-assisted development and engineering workflows.
The work is supported by Start-up Fund for RAPs under the Strategic Hiring Scheme of PolyU (P0063346).
\fi

\subsection*{Reproducibility Statement}

The experiments use Ascend NPUs.
Section~\ref{sec:experiments} records the evaluation protocols, metrics, and controlled training settings.
Appendix~\ref{app:pipeline-adaptations} specifies the model adaptations, while Appendix~\ref{app:geometry-supervision} and Appendix~\ref{app:semantic-supervision} detail pseudo-target construction and supervision.

\bibliography{iclr2027_conference}
\bibliographystyle{iclr2027_conference}

\clearpage
\appendix
\section*{Appendix}
Here we provide supplementary material to support the main paper:
\begin{flushleft}
\hspace{2em}\textbf{A.}\quad Additional Experiments.\\
\hspace{4em}\textbf{A.1.}\quad DReSS-D Results.\\
\hspace{4em}\textbf{A.2.}\quad KITTI-CVL Results.\\
\hspace{4em}\textbf{A.3.}\quad Qualitative Results.\\
\hspace{4em}\textbf{A.4.}\quad GT-Aligned BEV Correspondence Localization.\\
\hspace{2em}\textbf{B.}\quad Implementation Details and Pipeline Adaptations.\\
\hspace{4em}\textbf{B.1.}\quad Shared Implementation Details.\\
\hspace{4em}\textbf{B.2.}\quad Coordinate and Sampling Conventions.\\
\hspace{4em}\textbf{B.3.}\quad Geometry Supervision Details.\\
\hspace{4em}\textbf{B.4.}\quad Semantic Supervision Details.\\
\hspace{4em}\textbf{B.5.}\quad Training and Evaluation Configuration.\\
\hspace{4em}\textbf{B.6.}\quad Pipeline Adaptations.\\
\hspace{6em}\textbf{B.6.1.}\quad GS-FG$^2$.\\
\hspace{6em}\textbf{B.6.2.}\quad GS-ViewBridge.\\
\hspace{6em}\textbf{B.6.3.}\quad GS-DenseFlow.
\end{flushleft}
\section{Additional Experiments}
\label{app:additional-experiments}
\subsection{DReSS-D Results}
\label{app:dressd-results}

Table~\ref{tab:dressd} reports the DReSS-D unknown-orientation regimes.
Published CCVPE and ViewBridge values follow the DReSS-D comparison in ViewBridge~\citep{xia2025viewbridge}, which reports this benchmark without RANSAC refinement.
We therefore evaluate GS-ViewBridge, the reproduced FG$^{2\dagger}$ baseline, and GS-FG$^2$ without RANSAC refinement on the official DReSS-D test splits, so that every entry in the table is directly comparable.

\begin{table}[H]
\centering
\caption{DReSS-D test results under unknown orientation.
All results are reported without RANSAC refinement, following the published DReSS-D comparison.
CCVPE and ViewBridge values follow that comparison, and FG$^{2\dagger}$ denotes our re-implementation.
Best, second-best, and third-best distinct values within each metric are marked in red bold, blue bold, and black bold, respectively; ties share the same rank.
Percentages in parentheses after every metric report the relative change from the corresponding base model, computed as $(E_{\mathrm{GS}}-E_{\mathrm{base}})/E_{\mathrm{base}}\times100\%$.}
\label{tab:dressd}
\resizebox{\linewidth}{!}{%
\begin{tabular}{l cc cc cc cc}
\toprule
\multirow{2}{*}{Method} & \multicolumn{4}{c}{Cross-area} & \multicolumn{4}{c}{Same-area}\\
\cmidrule(lr){2-5}\cmidrule(lr){6-9}
& \multicolumn{2}{c}{Localization (m)$\downarrow$} & \multicolumn{2}{c}{Orientation ($^\circ$)$\downarrow$} & \multicolumn{2}{c}{Localization (m)$\downarrow$} & \multicolumn{2}{c}{Orientation ($^\circ$)$\downarrow$}\\
& Mean & Median & Mean & Median & Mean & Median & Mean & Median\\
\midrule
CCVPE & 6.050 & \textcolor{red}{\textbf{2.230}} & 37.390 & 10.270 & 3.010 & \textcolor{red}{\textbf{1.020}} & 14.440 & 7.940 \\
FG$^{2\dagger}$ & 6.422 & 3.814 & 16.466 & \textcolor{blue}{\textbf{2.411}} & 5.508 & 2.850 & 14.124 & 2.129 \\
\textcolor{black}{\textbf{GS-FG$^{\boldsymbol{2}}$}} & \textcolor{black}{\textbf{5.926}}\,(-7.72\%) & 3.396\,(-10.96\%) & \textcolor{black}{\textbf{14.696}}\,(-10.75\%) & 2.234\,(-7.34\%) & \textcolor{black}{\textbf{4.886}}\,(-11.29\%) & 2.475\,(-13.16\%) & \textcolor{black}{\textbf{11.921}}\,(-15.60\%) & \textcolor{black}{\textbf{1.939}}\,(-8.92\%) \\
ViewBridge & \textcolor{blue}{\textbf{4.220}} & \textcolor{blue}{\textbf{2.370}} & \textcolor{blue}{\textbf{12.580}} & \textcolor{black}{\textbf{2.600}} & \textcolor{red}{\textbf{2.940}} & \textcolor{blue}{\textbf{1.490}} & \textcolor{blue}{\textbf{8.470}} & \textcolor{red}{\textbf{1.620}} \\
\textcolor{black}{\textbf{GS-ViewBridge}} & \textcolor{red}{\textbf{4.108}}\,(-2.65\%) & \textcolor{black}{\textbf{2.468}}\,(+4.14\%) & \textcolor{red}{\textbf{8.810}}\,(-29.97\%) & \textcolor{red}{\textbf{1.976}}\,(-24.00\%) & \textcolor{blue}{\textbf{3.186}}\,(+8.37\%) & \textcolor{black}{\textbf{1.707}}\,(+14.56\%) & \textcolor{red}{\textbf{6.854}}\,(-19.08\%) & \textcolor{black}{\textbf{1.668}}\,(+2.96\%) \\
\bottomrule
\end{tabular}}
\end{table}

The DReSS-D results show that GeoSem-BEV's effect depends on the underlying framework.
GS-FG$^2$ reduces both mean localization and orientation errors in every reported same-area and cross-area setting.
Under unknown orientation, its localization errors decrease by 7.7\% and 11.3\%, and its orientation errors by 10.8\% and 15.6\%, on cross-area and same-area, respectively.
GS-ViewBridge reduces mean orientation error under unknown orientation by 30.0\% on cross-area and 19.1\% on same-area, and slightly reduces cross-area localization error, while its same-area localization error increases.
These results indicate that the constraints can reduce localization ambiguity, while the magnitude and type of improvement vary with the difficulty of the evaluated samples and the matching properties of the base framework.

We next evaluate KITTI-CVL to examine whether these gains transfer from panoramic ground observations to limited-field-of-view imagery.
\subsection{KITTI-CVL Results}
\label{app:kitti-results}

The KITTI-CVL benchmark evaluates satellite-ground localization from limited-field-of-view ground observations and satellite references~\citep{shi2022cvlnet}.
We include it to assess whether GeoSem-BEV transfers beyond panoramic ground views while retaining competitive localization and orientation performance.
Table~\ref{tab:kitti-main} reports the KITTI-CVL comparison, with published baseline values collected in Loc$^2$~\citep{xia2026loc}.
The GS-FG$^2$ cross-area and same-area $\pm10^\circ$ results and the GS-ViewBridge results are evaluated on the official test split without RANSAC.

\begin{table}[t]
\centering
\caption{KITTI-CVL test results.
Orientation noise is sampled uniformly within $\pm10^\circ$ during training and testing.
Lateral and longitudinal recalls are omitted.
Best, second-best, and third-best distinct values within each area and orientation setting are marked in red bold, blue bold, and black bold, respectively; ties share the same rank.
For GS-FG$^2$ and GS-ViewBridge, percentages in parentheses after every metric report relative changes from the corresponding base model.
A dash denotes an unreported source metric.}
\label{tab:kitti-main}
\resizebox{\linewidth}{!}{%
\begin{tabular}{cllcccccc}
\toprule
Area & Ori. & Method & \multicolumn{2}{c}{Loc. (m)$\downarrow$} & \multicolumn{2}{c}{Ori. ($^\circ$)$\downarrow$} & \multicolumn{2}{c}{Ori. (\%)$\uparrow$}\\
\cmidrule(lr){4-5}\cmidrule(lr){6-7}\cmidrule(lr){8-9}
 & & & Mean & Median & Mean & Median & R@1$^\circ$ & R@5$^\circ$\\
\midrule
\multirow{8}{*}{\rotatebox{90}{Cross-area}} & \multirow{8}{*}{$\pm10^\circ$} & GGCVT & -- & -- & -- & -- & \textcolor{red}{\textbf{98.98}} & \textcolor{red}{\textbf{100.00}} \\
 &  & CCVPE & 9.160 & \textcolor{blue}{\textbf{3.330}} & \textcolor{red}{\textbf{1.550}} & \textcolor{red}{\textbf{0.840}} & \textcolor{blue}{\textbf{57.72}} & \textcolor{blue}{\textbf{96.19}} \\
 &  & HC-Net & 8.470 & 4.570 & \textcolor{black}{\textbf{3.220}} & \textcolor{black}{\textbf{1.630}} & 33.58 & 83.78 \\
 &  & FG$^2$ & \textcolor{black}{\textbf{7.310}} & 4.150 & 3.620 & 2.370 & 23.03 & 77.84 \\
 &  & \textcolor{black}{\textbf{GS-FG$^{\boldsymbol{2}}$}} & \textcolor{black}{\textbf{7.026}}\,(-3.89\%) & 4.372\,(+5.35\%) & 3.485\,(-3.73\%) & 2.003\,(-15.49\%) & 28.07\,(+21.88\%) & 80.48\,(+3.39\%)\\
 &  & Loc$^2$ & \textcolor{red}{\textbf{5.600}} & \textcolor{blue}{\textbf{3.010}} & 3.320 & 2.120 & 26.03 & 80.68 \\
 &  & ViewBridge & 6.700 & 3.270 & 3.150 & 1.990 & 28.62 & 81.63\\
 &  & \textcolor{black}{\textbf{GS-ViewBridge}} & \textcolor{blue}{\textbf{5.880}}\,(-12.24\%) & \textcolor{red}{\textbf{2.686}}\,(-17.86\%) & 3.635\,(+15.40\%) & 1.732\,(-12.96\%) & 30.64\,(+7.06\%) & \textcolor{black}{\textbf{85.60}}\,(+4.86\%)\\
\midrule
\multirow{8}{*}{\rotatebox{90}{Same-area}} & \multirow{8}{*}{$\pm10^\circ$} & GGCVT & -- & -- & -- & -- & \textcolor{red}{\textbf{99.10}} & \textcolor{red}{\textbf{100.00}} \\
 &  & CCVPE & 1.220 & 0.620 & \textcolor{black}{\textbf{0.670}} & \textcolor{black}{\textbf{0.540}} & 77.39 & {\textbf{99.95}} \\
 &  & HC-Net & \textcolor{black}{\textbf{0.800}} & \textcolor{black}{\textbf{0.500}} & \textcolor{red}{\textbf{0.450}} & \textcolor{blue}{\textbf{0.330}} & \textcolor{blue}{\textbf{91.35}} & \textcolor{black}{99.84} \\
 &  & FG$^2$ & \textcolor{blue}{\textbf{0.750}} & 0.510 & 0.930 & 0.660 & 67.27 & 98.91 \\
 &  & \textcolor{black}{\textbf{GS-FG$^{\boldsymbol{2}}$}} & 0.830\,(+10.67\%) & 0.582\,(+14.12\%) & 0.795\,(-14.52\%) & 0.590\,(-10.61\%) & 72.65\,(+7.99\%) & 99.55\,(+0.65\%)\\
 &  & Loc$^2$ & 1.130 & 0.770 & 1.970 & 1.430 & 36.68 & 92.84 \\
 &  & ViewBridge & \textcolor{red}{\textbf{0.680}} & \textcolor{red}{\textbf{0.460}} & 1.100 & 0.720 & 61.77 & 97.97\\
 &  & \textcolor{black}{\textbf{GS-ViewBridge}} & \textcolor{black}{\textbf{0.796}}\,(+17.06\%) & 0.539\,(+17.17\%) & 0.805\,(-26.82\%) & 0.566\,(-21.39\%) & 74.08\,(+19.93\%) & \textcolor{blue}{\textbf{99.98}}\,(+2.05\%)\\
\bottomrule
\end{tabular}}
\end{table}

KITTI-CVL tests whether GeoSem-BEV transfers to limited-field-of-view ground images.
In cross-area evaluation, GS-FG$^2$ reduces both mean localization and orientation errors, while GS-ViewBridge reduces mean localization error.
In same-area evaluation, both implementations reduce mean orientation error and improve orientation recall, but their mean localization errors increase.
These results support the applicability of GeoSem-BEV to limited-field-of-view scenes and suggest that its constraints can reduce descriptor matching ambiguity, particularly for orientation estimation.
However, improved orientation estimates do not guarantee lower localization error, and the localization effect varies with the underlying framework and area split.

\subsection{Qualitative Results}

\begin{figure}[t]
  \centering
  \includegraphics[width=\linewidth]{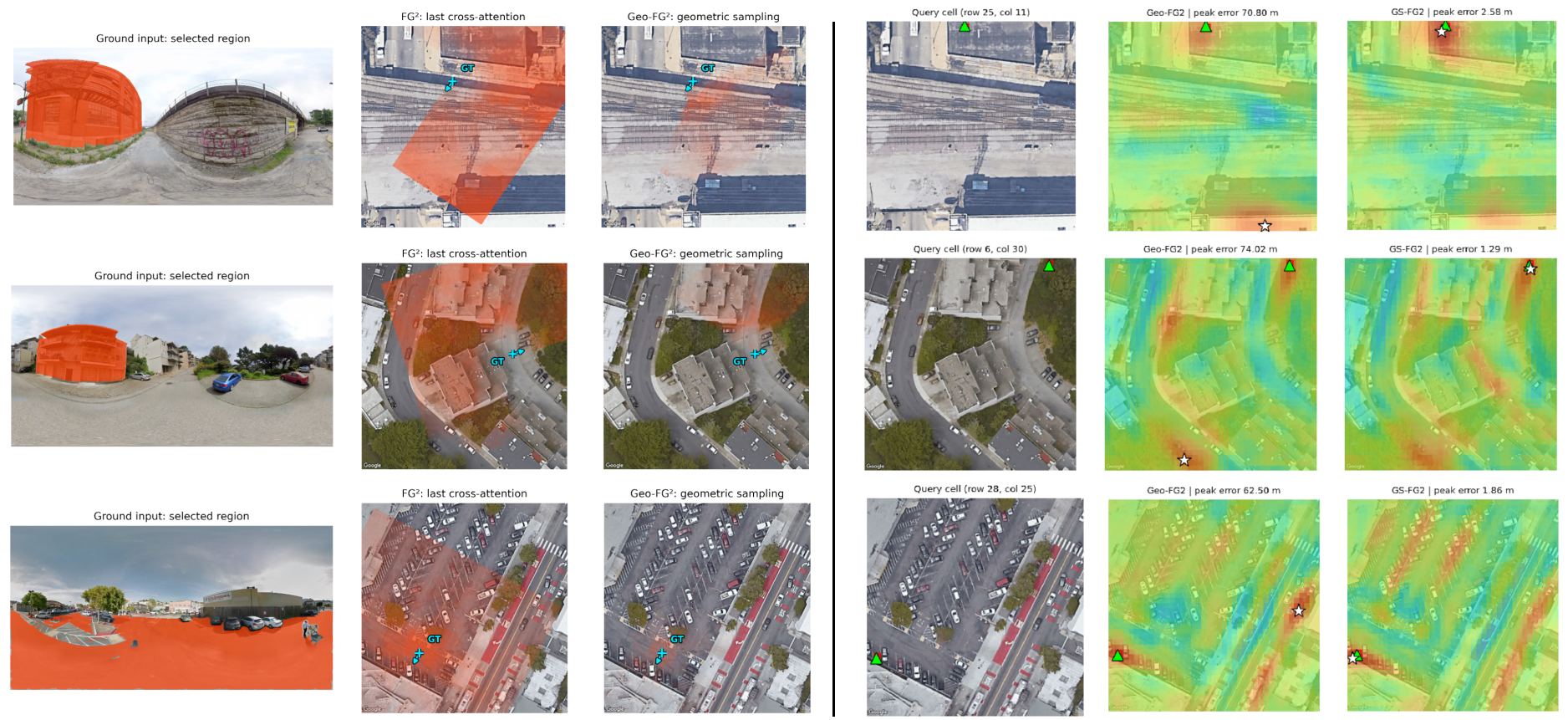}
  \caption{Qualitative visualization of geometry-constrained feature placement and explicitly supervised descriptor learning.
  In the left block, the orange overlay is the RGB content sampled from the selected ground-image region and projected onto the satellite images; the two columns of satellite images show the projection produced by FG$^{2\dagger}$ and Geo-FG$^2$, respectively.
  The cyan $+$ and arrow mark the ground-truth camera position and orientation used for alignment.
  In the right block, each heatmap shows the normalized descriptor-matching response for one ground-view BEV query over satellite locations: warmer colors indicate higher matching scores.
  The green triangle marks the ground-truth satellite location corresponding to the query cell, and the white star marks the highest-scoring predicted location; the reported peak error is the distance between these two markers.
  }
  \label{fig:geo-sem-qualitative}
\end{figure}

Figure~\ref{fig:geo-sem-qualitative} provides qualitative examples of the two effects targeted by our supervision.
The examples are selected from the VIGOR cross-area unknown-orientation evaluation cases used for the diagnostic visualizations.
In the feature-placement examples, adding radial-depth and vertical-height supervision produces projected ground-view content that is more spatially concentrated and better aligned with the corresponding structures in the satellite reference than the unconstrained baseline.
In the descriptor examples, explicit semantic supervision concentrates the matching response around the ground-truth location and suppresses responses from visually similar regions at different locations.
These observations illustrate how the geometric and semantic constraints address the two forms of localization ambiguity discussed in the main paper.

\begin{figure}[t]
  \centering
  \includegraphics[width=\linewidth]{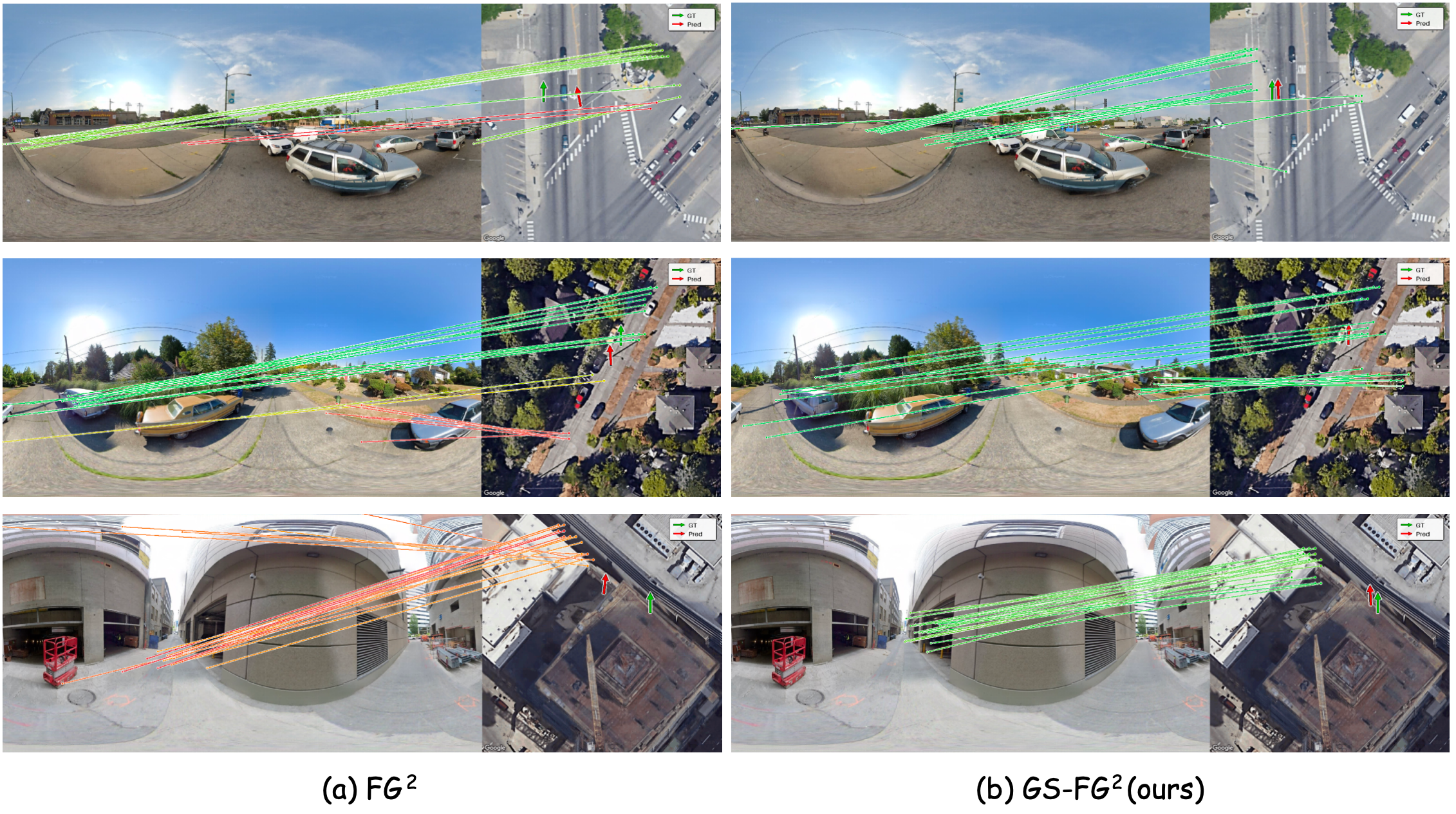}
  \caption{Qualitative correspondence comparison on challenging VIGOR samples.
  Each row shows the same ground--satellite image pair for FG$^2$ (left) and GS-FG$^2$ (right).
  Lines visualize selected correspondences; green and red arrows indicate the ground-truth and predicted camera poses, respectively.}
  \label{fig:correspondence-comparison}
\end{figure}

Figure~\ref{fig:correspondence-comparison} further compares the correspondence patterns produced by FG$^2$ and GS-FG$^2$ on challenging VIGOR samples.
In cases where FG$^2$ produces dispersed or spatially inconsistent correspondences, GS-FG$^2$ yields more coherent matches and pose estimates closer to the ground truth.
These examples illustrate that incorporating GeoSem-BEV can improve correspondence matching on samples that remain difficult for the original model.

\subsection{GT-Aligned BEV Correspondence Localization}

To directly evaluate whether radial geometry improves the spatial placement of ground-view BEV features, we perform a post-hoc correspondence-localization diagnostic on a fixed 5,000-sample subset of the VIGOR cross-area validation split with unknown orientation.
For each ground-view BEV cell whose ground-truth transformed position lies inside the satellite-view BEV, the ground-truth pose defines its target satellite location.
The predicted location is the satellite cell with the highest descriptor-matching score.
We report the Euclidean error between these locations and the proportion of valid correspondences within 1, 3, and 5 metres.
FG$^{2\dagger}$ and Geo-FG$^2$ use the same samples, descriptor matcher, and evaluation procedure to assess the geometry-constrained variant.

\begin{table}[h]
\centering
\caption{Ground-truth-aligned BEV correspondence localization on 5,000 VIGOR cross-area validation samples with unknown orientation.
Correspondence errors are measured from the highest-scoring satellite-view BEV cell to the aligned location.
Recalls are percentages.
Best values are in bold.}
\label{tab:bev-correspondence-localization}
\resizebox{\linewidth}{!}{%
\begin{tabular}{lcccccc}
\toprule
Method & Valid-cell ratio & Corr. error mean $\downarrow$ (m) & Corr. error median $\downarrow$ (m) & Corr.@1m $\uparrow$ & Corr.@3m $\uparrow$ & Corr.@5m $\uparrow$\\
\midrule
FG$^{2\dagger}$ & 0.726 & 9.095 & 3.662 & 9.29 & 41.20 & 60.40 \\
Geo-FG$^2$ & 0.726 & \textbf{8.427} & \textbf{3.244} & \textbf{10.64} & \textbf{45.52} & \textbf{64.86} \\
\bottomrule
\end{tabular}}
\end{table}

Geo-FG$^2$ reduces the mean correspondence error from 9.095 m to 8.427 m and the median error from 3.662 m to 3.244 m.
It also increases Corr.@1m, Corr.@3m, and Corr.@5m across all three thresholds.
A paired sample-level bootstrap gives a mean-error difference of $-0.669$ m with a 95\% confidence interval of $[-0.796,-0.539]$ m.
These results support that radial geometry places ground-view features closer to their ground-truth-aligned BEV correspondence locations.


%
\section{Implementation Details and Pipeline Adaptations}
\label{app:implementation-details}

\subsection{Shared Implementation Details}

The GS variants share radial geometric supervision, explicit semantic descriptor supervision, and same-class hard negatives.
Their feature encoders, aggregation normalization, and correspondence estimators differ.
The shared training and evaluation settings are summarized in Appendix~\ref{app:training-configuration}; implementation-specific departures are described below.

\subsection{Coordinate and Sampling Conventions}
\label{app:coordinate-conventions}

The ground-view query grid is indexed by two horizontal coordinates and one height coordinate.
A BEV grid contains horizontal cells; each cell has a column of 3D queries at different heights.
Satellite pixel offsets are converted to metres using the ground sampling distance, accounting for image resizing and feature-grid resolution.
In GS-FG$^2$, each horizontal axis spans $[-35.5,35.5]$ m with 41 samples, and the 11 heights span $[-10,10]$ m relative to the camera.
A horizontal metric coordinate $x$ maps to grid coordinate $(x+35.5)40/71$.
Equirectangular image sampling wraps horizontally.
Satellite-view sampling uses bilinear interpolation after resizing the satellite feature map to the $41\times41$ BEV resolution.
The main-text pose convention maps ground coordinates to satellite coordinates; the GS-FG$^2$ implementation internally uses row--column coordinates and the inverse satellite-to-ground transform.
Coordinate ordering and transform direction are converted consistently at the interface.

\subsection{Geometry Supervision Details}
\label{app:geometry-supervision}

The following target construction and height loss specify GS-FG$^2$; the separately supervised vertical distributions used by GS-ViewBridge and GS-DenseFlow are identified in Appendices~\ref{app:gs-viewbridge} and~\ref{app:gs-denseflow}.
The frozen ground-view SphereViT teacher from DA$^2$~\citep{li20252} predicts relative radial depth.
Its scale is set by dividing the assumed camera height of $2.5$ m by the 95th percentile of downward vertical distances in the lower image region.
The resulting metric pseudo-depth supervises surface presence, the 64-bin radial distribution over $0$--$35$ m, expected radial depth, and uncertainty.
The teacher uses valid downward rays below $0.55$ of the image height, extending to all valid downward rays when fewer than 32 candidates remain.

Depth Anything V2 predictions for the satellite view~\citep{yang2024depth} are min--max normalized and bilinearly resized to $41\times41$.
A normalized value $h\in[0,1]$ maps to continuous height-bin index $10h$, from which we construct a normalized Gaussian soft target $h^S_{ijk}$ with standard deviation $0.75$ bins.
This is a relative-height pseudo-target over the query bins, not a measured metric elevation.
The ground-truth pose $T^\ast$ aligns this target to the ground-view BEV grid by bilinear sampling, yielding $\hat h^G_{ijk}$.
For valid overlapping cells $\Omega$, the height loss is normalized by $\log 11$, the entropy of a uniform distribution over the 11 height candidates.
\begin{equation}
  \mathcal L_{\mathrm{height}}=
  -\frac{1}{|\Omega|\log 11}
  \sum_{(i,j)\in\Omega}\sum_{k=1}^{11}
  \hat h^G_{ijk}\log\!\left(\max(w_{ijk},\epsilon)\right),
  \qquad \epsilon=10^{-8}.
  \label{eq:height-loss}
\end{equation}
The target index $k$ corresponds to the same 11 height candidates used in Equation~\ref{eq:height-aggregation}.
Its gradient therefore updates the surface and radial predictions that generate $w_{ijk}$.

\subsection{Semantic Supervision Details}
\label{app:semantic-supervision}

For view $v\in\{G,S\}$, $Y_v$ and $P_v$ denote the soft semantic targets and predicted class probabilities.
Let $\mathcal V_v$ contain the valid BEV cells and $\mathcal C_v$ the classes with nonzero target mass.
With cell index $n$ and class index $c$, the macro-balanced soft cross-entropy is
\begin{equation}
  \mathcal L_{\mathrm{CE}}^{v}=
  \frac{1}{|\mathcal C_v|}
  \sum_{c\in\mathcal C_v}
  \frac{-\sum_{n\in\mathcal V_v}Y_{v,nc}\log P_{v,nc}}
  {\sum_{n\in\mathcal V_v}Y_{v,nc}},
  \qquad v\in\{G,S\}.
  \label{eq:semantic-ce}
\end{equation}
The ground-truth pose $T^\ast$ bilinearly warps $P_G$ to the satellite grid, producing $P_G^{T^\ast}$ over the valid overlap $\mathcal V_{T^\ast}$.
With $M_n=(P_{G,n}^{T^\ast}+P_{S,n})/2$, the consistency term is
\begin{equation}
  \mathcal L_{\mathrm{JS}}(P_G^{T^\ast},P_S)=
  \frac{1}{|\mathcal V_{T^\ast}|}
  \sum_{n\in\mathcal V_{T^\ast}}
  \frac{1}{2}\left[
  D_{\mathrm{KL}}(P_{G,n}^{T^\ast}\|M_n)+
  D_{\mathrm{KL}}(P_{S,n}\|M_n)
  \right].
  \label{eq:semantic-js}
\end{equation}

For direction $X\rightarrow Y$, with $(X,Y)\in\{(G,S),(S,G)\}$, let $\mathcal A_{X\rightarrow Y}$ contain the valid anchors.
The positive logit $s_a^+$ is defined by the bilinear soft positive, and $s_{a,n}^-$ denotes the matching logit of negative candidate $n$.
The set $\mathcal N_a$ contains the top 32 same-class candidates outside the one-cell neighbourhood.
The anchor weight $\alpha_a$ is the product of DA2-derived visibility and semantic confidence.
The directional loss is
\begin{equation}
  \mathcal L_{X\rightarrow Y}^{\mathrm{HN}}=
  \frac{
  \sum_{a\in\mathcal A_{X\rightarrow Y}}\alpha_a
  \left[
  \log\left(e^{s_a^+}+\sum_{n\in\mathcal N_a}e^{s_{a,n}^-}\right)-s_a^+
  \right]}
  {\sum_{a\in\mathcal A_{X\rightarrow Y}}\alpha_a},
  \qquad X,Y\in\{G,S\}.
  \label{eq:directional-hard-negative}
\end{equation}
Implementation-specific candidate confidence thresholds and back-end logits are given in Appendices~\ref{app:gs-fg2}--\ref{app:gs-denseflow}.

\subsection{Training and Evaluation Configuration}
\label{app:training-configuration}

The controlled GS-FG$^2$ experiments use frozen DINOv2 features~\citep{oquab2023dinov2} with 1024 channels, a $41\times41\times11$ query grid, 64 radial depth bins, and 128-dimensional descriptors.
The pose estimator samples 1024 correspondences.
Training uses Adam~\citep{kingma2014adam} with learning rate $10^{-4}$ and zero weight decay on Ascend NPUs.
Feature computation uses BF16; matching probabilities, losses, and pose solving use FP32.
The main VIGOR cross-area unknown-orientation experiment uses 25 epochs, seed 0, global batch size 24, and random rotations within $[-180^\circ,180^\circ]$.
Evaluation uses the official test split of 53{,}694 samples.
Depth predictions and SAM3 masks are precomputed or cached, and ground-truth poses align cross-view training targets.
GS-FG$^2$ and its ablations use the same RANSAC protocol unless stated otherwise.

\subsection{Pipeline Adaptations}
\label{app:pipeline-adaptations}

The following descriptions specify the evaluated implementations rather than asserting identical architectures across back ends.
We first describe GS-FG$^2$, the baseline used for the detailed supervision analysis, followed by GS-ViewBridge and GS-DenseFlow.

\subsubsection{GS-FG$^2$}
\label{app:gs-fg2}

GS-FG$^2$ retains the FG$^2$ descriptor projection, dual-softmax matching with a dustbin, and correspondence-based pose solver~\citep{xia2025fg2}.
Frozen DINOv2 features are sampled once on a $41\times41\times11$ grid spanning $[-35.5,35.5]$ m horizontally and $[-10,10]$ m vertically.
Unlike GS-ViewBridge, this implementation has no independent vertical head: $w_{ijk}=\ell_{ijk}$.
Satellite height targets directly supervise these likelihood weights.
The aggregated appearance features pass through six planar self-attention/FFN blocks and a 128-dimensional projector.
No radial-mass gate is applied after these blocks or to the matching coupling.

The shared semantic head supervises the final normalized descriptors, and same-class hard negatives act on the descriptor matching logits.
The base objective is $\mathcal L_{\mathrm{VCE}}+\beta\mathcal L_{\mathrm{InfoNCE}}$, with unit coefficients on the three added objectives.
The VIGOR unknown-orientation configuration uses $\beta=100$.
The RANSAC and direct Weighted Procrustes evaluations are reported separately.

\subsubsection{GS-ViewBridge}
\label{app:gs-viewbridge}

GS-ViewBridge retains the frozen DINOv2 encoder, satellite projector, planar context modules, similarity refinement, adaptive dustbin, and soft-coupling back end of the ViewBridge implementation~\citep{xia2025viewbridge}.
The ground branch replaces learned panorama cross-attention with analytic sampling at a $41\times41\times11$ metric grid.
Ground block11 and block23 features feed the radial head; block23 provides the sampled appearance.
The radial head predicts a surface probability, a 64-bin distribution over $0$--$35$ m, and uncertainty.

A separate vertical head predicts $v_{ijk}$, normalized across the 11 heights of each BEV column.
With the radial likelihood $\ell_{ijk}$ from Equation~\ref{eq:radial-weight}, the aggregation weight is $w_{ijk}=v_{ijk}\ell_{ijk}$.
The joint weights are not renormalized over height.
Their sum $m_{ij}=\sum_k w_{ijk}$ also gates the ground features before and after the six planar context blocks.
Satellite height supervision acts on $v_{ijk}$.
The resulting 128-dimensional descriptors enter ViewBridge's similarity refinement and adaptive dustbin.
Ground mass additionally weights correspondence sampling and alignment.
The model samples 1024 pairs for Weighted Procrustes; RANSAC is enabled only for the corresponding evaluation protocol.

The shared semantic head is LayerNorm--Linear($128,64$)--GELU--Linear($64,3$).
The hard-negative objective reads the refined matching logits, uses bilinear soft positives, excludes a Chebyshev radius of one cell, and selects 32 same-class candidates with class confidence at least $0.5$.
The base objective combines the pose loss and the existing InfoNCE loss; the geometry, semantic, and hard-negative terms have unit coefficients.
On VIGOR and DReSS-D, the pose-loss coefficient is $1$, and the InfoNCE coefficient is $100$ for unknown orientation and $1$ for known orientation.

\subsubsection{GS-DenseFlow}
\label{app:gs-denseflow}

GS-DenseFlow constructs geometry-constrained ground BEV representations using ResNet18 features~\citep{he2016deep} and retains the RAFT-based flow estimator~\citep{teed2020raft} used by DenseFlow~\citep{song2023learning}.
The adapted builder samples ground features at 3D queries, predicts radial and vertical distributions, and forms their product $j_{ijk}=\ell_{ijk}v_{ijk}$.
It normalizes the appearance aggregation by the column mass: $w_{ijk}=j_{ijk}/\max(\sum_k j_{ijk},\epsilon)$, with $\epsilon=10^{-6}$.
The unnormalized mass is retained separately as a geometric confidence signal.
Thus, this implementation uses the same geometric assignment principle but a different aggregation normalization from GS-ViewBridge and GS-FG$^2$.

Semantic supervision is applied to the actual ground and satellite feature maps used by RAFT to construct correlations.
Its shared head maps 256-dimensional descriptors through a 64-dimensional hidden layer to three classes.
Same-class hard negatives are selected from the actual level-zero correlation field.
The flow estimator and confidence prediction remain in the pipeline, and geometric mass weights the confidence used for pose alignment.
The original DenseFlow training objective is retained, with geometry, semantic, and hard-negative terms added at unit weight.

\end{document}